\documentclass[10pt,lettersize,journal]{IEEEtran}

\usepackage{amsmath,amssymb,amsfonts}

\usepackage{algorithm}        
\usepackage{algpseudocode}    

\usepackage{graphicx}
\usepackage{array}
\usepackage{booktabs}
\usepackage{multirow}
\usepackage{longtable}
\usepackage{multicol}
\usepackage{subfig} 
\usepackage[T1]{fontenc}
\usepackage[utf8]{inputenc}  
\usepackage{microtype}
\usepackage{inconsolata}
\usepackage{textcomp}
\usepackage{stfloats}
\usepackage{url}
\usepackage{verbatim}
\usepackage{makecell}
\usepackage{tcolorbox}
\tcbuselibrary{listingsutf8}
\usepackage{listings}         

\usepackage{cite}
\usepackage[hidelinks]{hyperref}

\begin{document}
\title{Unlocking Multi-View Insights in Knowledge-Dense Retrieval-Augmented Generation}

\author{Guanhua Chen, Wenhan Yu, Xiao Lu, Xiao Zhang, Erli Meng, and Lei Sha*%
\thanks{*~Corresponding author}%
\thanks{Guanhua Chen, Wenhan Yu, and Lei Sha are with the School of Artificial Intelligence, Beihang University, China. 
Lei Sha is also with Zhongguancun Laboratory, Beijing, China.}}

\maketitle

\begin{abstract}
  While Retrieval-Augmented Generation (RAG) plays a crucial role in the application of Large Language Models (LLMs), existing retrieval methods in knowledge-dense domains like law and medicine still suffer from the insufficient utilization of multi-perspective views embedded within domain-specific corpora, which are essential for improving interpretability and reliability. Previous research on multi-view retrieval often focused solely on different semantic forms of queries, neglecting the expression of specific domain knowledge perspectives. This paper introduces a novel multi-view RAG framework, \textit{MVRAG}, tailored for knowledge-dense domains, which leverages machine learning techniques for professional perspectives extraction and intention-aware query rewriting from multiple domain viewpoints to enhance retrieval precision, thereby improving the effectiveness of the final inference. Experiments conducted on both retrieval and generation tasks demonstrate substantial improvements in generation quality while maintaining retrieval performance in complex, knowledge-dense scenarios.
\end{abstract}


\section{Introduction}
\IEEEPARstart{I}{n} the ever-evolving domain of natural language processing,  retrieval-augmented generation (RAG) stands as a cornerstone technology that synergistically combines large language models (LLMs)  with a vast corpus of external knowledge~\cite{gao2023retrieval}. This integration endows RAG systems with a remarkable capacity for nuanced language understanding and sophisticated information retrieval. Such capabilities are especially transformative in knowledge-dense domains like \textit{law}, \textit{medicine} and \textit{biology}, where RAG not only augments contextual comprehension but also improves the interpretability and reliability of the models. Despite these advantages,  current RAG implementations face substantial challenges in adequately serving the complicated and multifaceted information requisites inherent in these specialized fields.
\begin{figure}[htbp]
  \centering
  \includegraphics[width=0.8\columnwidth,  page=1]{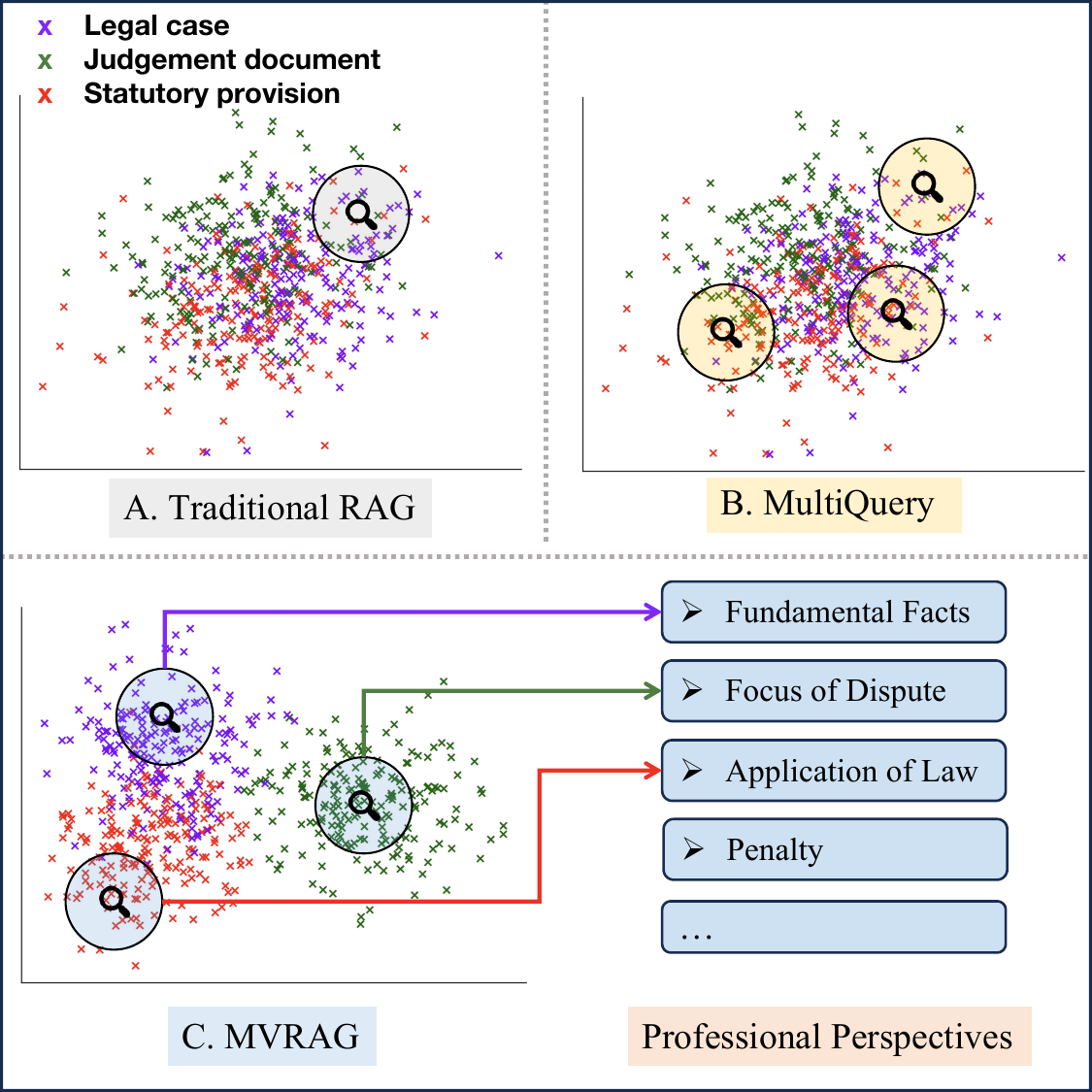} 
  \caption{A t-SNE visualization of retrieval results from different methods using a legal database with manually added category labels, displayed in different colors in the plot. The magnifying glass represents the query vector, while the circles represent the retrieval results. }
  \label{fig:view}
\end{figure}
Predominant among these challenges is the insufficient utilization of multi-view information embedded within domain-specific corpora. Unlike general-purpose corpora such as Wikipedia, domain-specific corpora contain self-organizing structural information, which is reflected in the distribution patterns of vectors within the vector space they form. These patterns represent the embedded \textit{professional perspectives} of the domain, guiding deeper and more thorough utilization of the corpus. These perspectives often play a more significant role than superficial textual similarity. In the legal domain, for example, the \textit{focus of dispute} between cases is more crucial than mere textual overlap; in medical scenarios, the relevance of \textit{medical history} or \textit{symptoms} is often more diagnostically valuable than literal similarity, which can be obscured by irrelevant information.

\begin{figure*}[htbp]
  \centering
  \includegraphics[width=0.8\textwidth,  page=1]{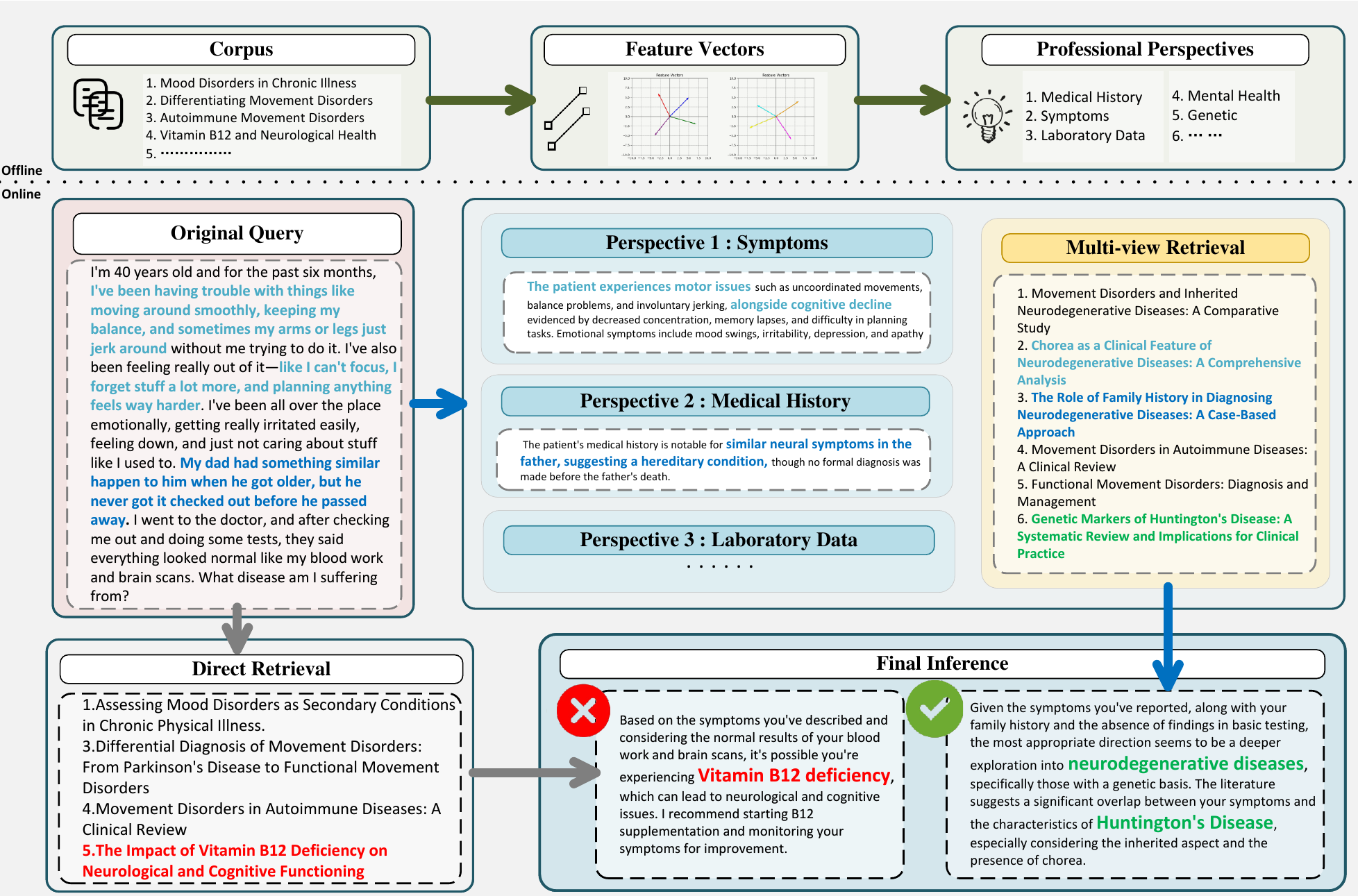} 
\caption{A case study showcasing the effectiveness of a multi-view retrieval framework in accurately diagnosing Huntington’s disease. The model corrects an initial misdiagnosis of Vitamin B12 deficiency by refining the search criteria to focus on neurodegenerative symptoms and family medical history, thus demonstrating the importance of multi-view search strategies in medical diagnostics. The professional perspectives used in the framework were determined during the offline part, ensuring their domain-specific relevance. The detailed case study is provided in Section~\ref{app:medical case}.}
  \label{fig:casestudy}
\end{figure*}

Traditional retrieval methods~\cite{gao2023retrieval}, as shown in Figure~\ref{fig:view}A, primarily rely on full-text similarity, retrieving only a limited subset of vectors around the query vector. This often overlooks essential professional nuances, leading to imprecise or incomplete results~\cite{wang2023query2doc}. To mitigate this, current multi-view retrieval research reformulates queries to adjust their position in the vector space (Figure~\ref{fig:view}B)~\cite{ma2023query}\cite{langchain2024multiquery}. Recent systems such as MC-indexing~\cite{dong2024mc} and NeuSym-RAG~\cite{cao2025neusym} pursue “multi-view” chiefly in a representational sense—indexing each section with raw text, keywords, and summaries or combining neural and symbolic backends—to broaden retrieval coverage.\cite{xu2025multigranularity} However, they do not decompose the query into domain-internal professional perspectives, which is crucial for capturing the multi-facet semantics required in knowledge-dense domains like law and medicine. Figure~\ref{fig:casestudy} highlights how the lack of multi-view insights can lead to erroneous outcomes, emphasizing the need for a more nuanced and context-aware retrieval approach.

To address these challenges and overcome the limitations of existing methods, we propose a novel multi-view retrieval framework, MVRAG. This system is built upon three key stages: 1) professional perspectives extraction, 2) intention recognition and query rewriting, and 3) retrieval augmentation. For a specialized corpus, we first use machine learning techniques like PCA to extract domain-specific perspectives. The query’s intention is identified by measuring its similarity to these perspectives, followed by rewriting the query for each perspective using an LLM. Results are retrieved and re-ranked based on perspective importance and retrieval similarity, then fed into the LLM generator. This process, illustrated in Figure~\ref{fig:view}C, distributes retrieval outcomes across multiple professional perspectives, enabling comprehensive corpus utilization.

We tested MVRAG on legal and medical qualification exams, where it significantly outperformed traditional RAG, especially on challenging multiple-choice and subjective questions. The contributions of this article mainly include the following three aspects:

\begin{itemize}
  \item We present MVRAG, a framework that integrates professional perspectives extraction, intention-aware multi-view query rewriting, and retrieval re-ranking to enable multi-perspective analysis in knowledge-intensive domains. This system is designed for easy integration and introduces a new paradigm for applying RAG in specialized fields.
  \item We propose an innovative approach that incorporates traditional machine learning methods into the RAG framework for feature extraction from domain-specific corpora. By combining these methods with query rewriting techniques, this approach enhances both retrieval accuracy and depth.
  \item We conducted experiments on both retrieval and generation tasks, highlighting the importance of multi-perspective strategies in knowledge-dense RAG tasks, validating their effectiveness and paving the way for future research.
\end{itemize}

\section{Related Work}
\subsection{Retrieval-Augmented Generation (RAG)}
\noindent Retrieval-augmented generation,  commonly known as RAG~\cite{izacard2022atlas, huo2023retrieving, guu2020retrieval, lewis2020retrieval} has become a prevalent technique to enhance LLMs. This approach integrates LLMs with retrieval systems,  enabling them to access domain-specific knowledge and base their responses on factual information~\cite{khattab2021relevance}. Additionally,  RAG provides a layer of transparency and interpretability by allowing these systems to cite their sources~\cite{shuster2021retrieval}.

Recent studies have demonstrated enhancements in the quality of output results across various scenarios through techniques such as appending documents retrieved by RAG to the inputs of LLMs,  and training unified embedded models. These methods have shown effective improvements in the performance of LLM outputs by leveraging RAG's ability to access and incorporate relevant information from external sources~\cite{ram2023context, es2023ragas}.

\subsection{Domain-Specific Large Language Models}
Due to the general nature of LLMs,  their expertise in specific domains such as \textit{law},  \textit{finance},  and \textit{healthcare} is often limited. Recent research focuses on enhancing domain-specific expertise through Knowledge Enhancement techniques,  introducing specific domain knowledge,  and employing innovative training methods to address the issue of hallucinations in models. This approach has become a key strategy for improving the professionalism of vertical domain large models~\cite{xi2023towards, yao2023knowledge, zhao2023survey, zhu2023large}.

In the legal field,  a number of well-known large language models have been born through methods such as secondary training,  instruction fine-tuning,  and RAG,  such as wisdomInterrogatory\footnote{\url{https://github.com/zhihaiLLM/wisdomInterrogatory}},  DISC-LawLLM\footnote{\url{https://github.com/FudanDISC/DISC-LawLLM}},  PKUlaw\footnote{\url{https://www.pkulaw.net/}},  and ChatLaw~\cite{cui2023chatlaw}. These models have been specifically designed to address the unique requirements of the legal domain,  providing specialized knowledge and expertise to enhance the quality and relevance of legal information retrieval and generation.

As for the medical domain,  the development of domain-specific LLMs has been a key focus in recent research. Models like Huatuo\footnote{\url{https://github.com/SCIR-HI/Huatuo-Llama-Med-Chinese}},  Zhongjing\footnote{\url{https://github.com/SupritYoung/Zhongjing}},  and Doctor-Dignity\footnote{\url{https://github.com/llSourcell/Doctor-Dignity}} have been specifically designed to cater to the complex and nuanced requirements of the medical field,  providing enhanced capabilities for medical information retrieval,  diagnosis,  and treatment planning. 
\begin{figure*}[htbp]
  \centering
  \includegraphics[width=0.9\textwidth,  page=1]{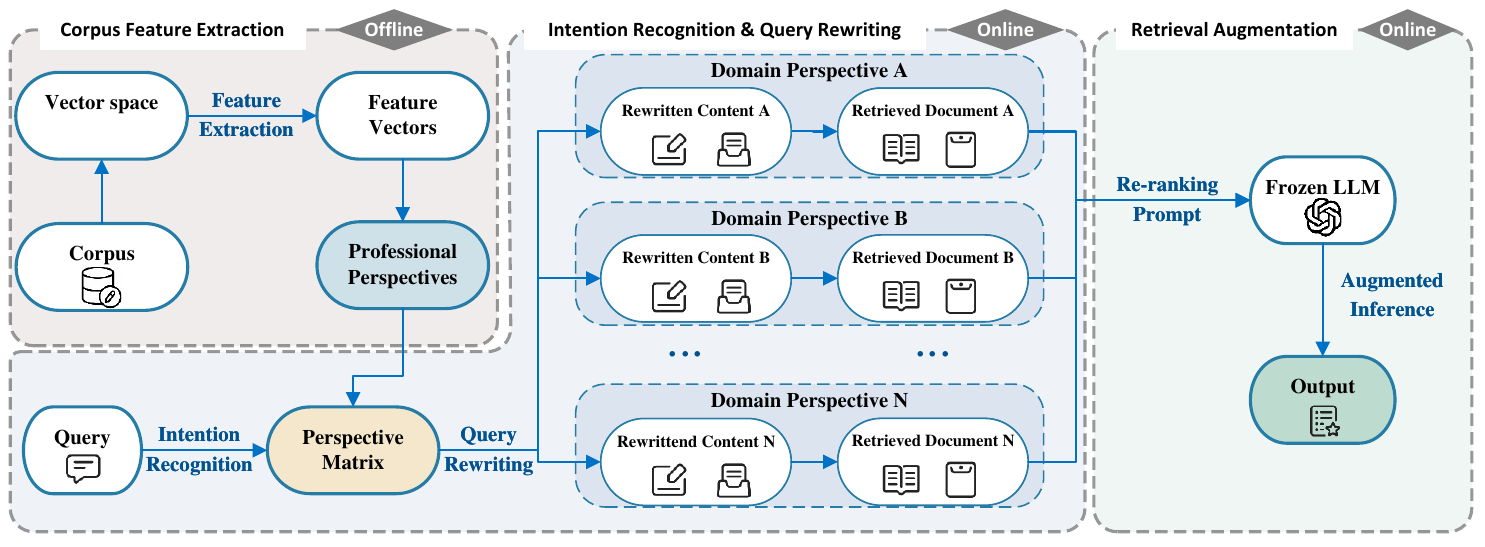} 
  \caption{Framework of our Multi-View RAG System. This figure demonstrates the system's core processes: Professional Perspectives Extraction, Intention Recognition and Query Rewriting,  and Retrieval Augmentation,  emphasizing the multi-view insights approach for intention-aware query rewriting}
  \label{fig:pdfimage}
  \end{figure*}
\subsection{Query Rewriting}
In RAG, user queries are often imprecise or under-specified, leading to inaccurate or unanswerable responses. 
Query rewriting mitigates this by refining queries to better align with document semantics, thereby improving retrieval accuracy. 
Gao et al.~\cite{gao2022precise} propose conjectural document embeddings to maximize query–document congruence, while Wang et al.~\cite{wang2023query2doc} leverage LLMs for query reformulation and pseudo-document generation, both showing notable gains in retrieval performance~\cite{ma2023query}. 
More recently, FLARE (Forward-Looking Active Retrieval Augmented Generation)~\cite{jiang2023flare} extends this idea by performing dense, iterative retrieval during generation, triggering retrieval whenever confidence is low.

\subsection{Multi-view and Hybrid Retrieval}
Recent efforts explore multi-view or hybrid retrieval. \cite{li2025mfar}
For example, MC-indexing~\cite{dong2024mc} introduces multi-view content-aware indexing for long documents, and NeuSym-RAG~\cite{cao2025neusym} integrates neural and symbolic retrieval with multi-view structuring for PDF QA. 
However, these methods focus on \emph{formal multi-view representations} (e.g., text, keywords, summaries), rather than decomposing queries into domain-specific perspectives. \cite{long2025diver}

Different from previous research,  our work concentrates on the multi-perspective professional information within specific domains,  which is a critical aspect often overlooked before. Instead of merely adjusting queries for semantic alignment,  our novel multi-view retrieval framework captures the complex relationships and local nuances inherent to each domain, enhancing the retrieval process by focusing on the multi-perspective aspects of domain-specific knowledge. 
\section{Method}
\subsection{Overview}

\noindent The Multi-View Retrieval-Augmented Generation (MVRAG) system comprises three main components: (1)~\textbf{Professional Perspectives Extraction}, (2)~\textbf{Intention Recognition \& Query Rewriting}, and (3)~\textbf{Retrieval Augmentation}.

\textbf{Professional Perspectives Extraction (Offline):} The domain-specific corpus is embedded into a vector space using a tokenizer model. Techniques such as Non-negative Matrix Factorization (NMF) and Principal Component Analysis (PCA) are applied to extract \textit{professional perspectives}, representing key viewpoints in the corpus. This offline step provides reusable domain representations for later stages.

\textbf{Intention Recognition \& Query Rewriting (Online):} A new query is vectorized and aligned with the extracted perspectives, producing a \textit{Perspective Vector} that reflects its relevance to each perspective. Based on this alignment, the query is rewritten for each relevant perspective using a rewriter model. Rewritten queries are used for similarity-based document retrieval.

\textbf{Retrieval Augmentation (Online):} Retrieved documents are re-ranked by combining similarity scores with perspective importance weights. The top-ranked documents are integrated into a structured prompt, enabling the LLM to generate a contextually rich, multi-perspective response.

This pipeline integrates offline preprocessing with real-time query handling, ensuring efficient and contextually nuanced responses. Figure~\ref{fig:pdfimage} illustrates the overall workflow.

\subsection{Professional Perspectives Extraction}

In detail, professional perspectives are topic terms that represent various aspects of a specific domain. These topic terms are then used to rewrite the query as is described in Section~\ref{sec:rewrite}. In this section, we first  capture the vector-level structural patterns in the corpus by PCA~(\textbf{Feature Vector Extraction}), and then we employ NMF to identify candidate lists of (word-level) perspective-related topic terms (\textbf{Topic Modeling}). Finally, we  translate the \textbf{feature vectors} into the  \textbf{topic terms} (\textbf{Alignment of Perspectives}).

\paragraph{Feature Vector Extraction.}
Each document \( d_i \in C \) in the corpus \( C \) is embedded as a vector \( \mathbf{v}_i \in \mathbb{R}^n \) in a high-dimensional vector space  generated using a domain-adapted embedding model that aligns with the LLM employed in downstream tasks. To uncover the underlying structure of the corpus and reduce dimensionality, Principal Component Analysis (PCA) \footnote{PCA was implemented using the \texttt{scikit-learn} library~\cite{scikit-learn}, version 1.6.0. For more details, see \url{https://scikit-learn.org/stable/}.} is applied:
\begin{equation}
\mathbf{v}_1, \mathbf{v}_2, \dots, \mathbf{v}_n = \text{PCA}(\mathbf{C}),
\end{equation}
where \( \mathbf{v}_i \) represents the \( i \)-th principal component, capturing key structural patterns of the corpus. The number of principal components \( n \) is determined by retaining 90\% of the cumulative variance, ensuring a balance between computational efficiency and the preservation of key information.

\begin{figure}[t]
  \centering
  \includegraphics[width=\columnwidth, page=1]{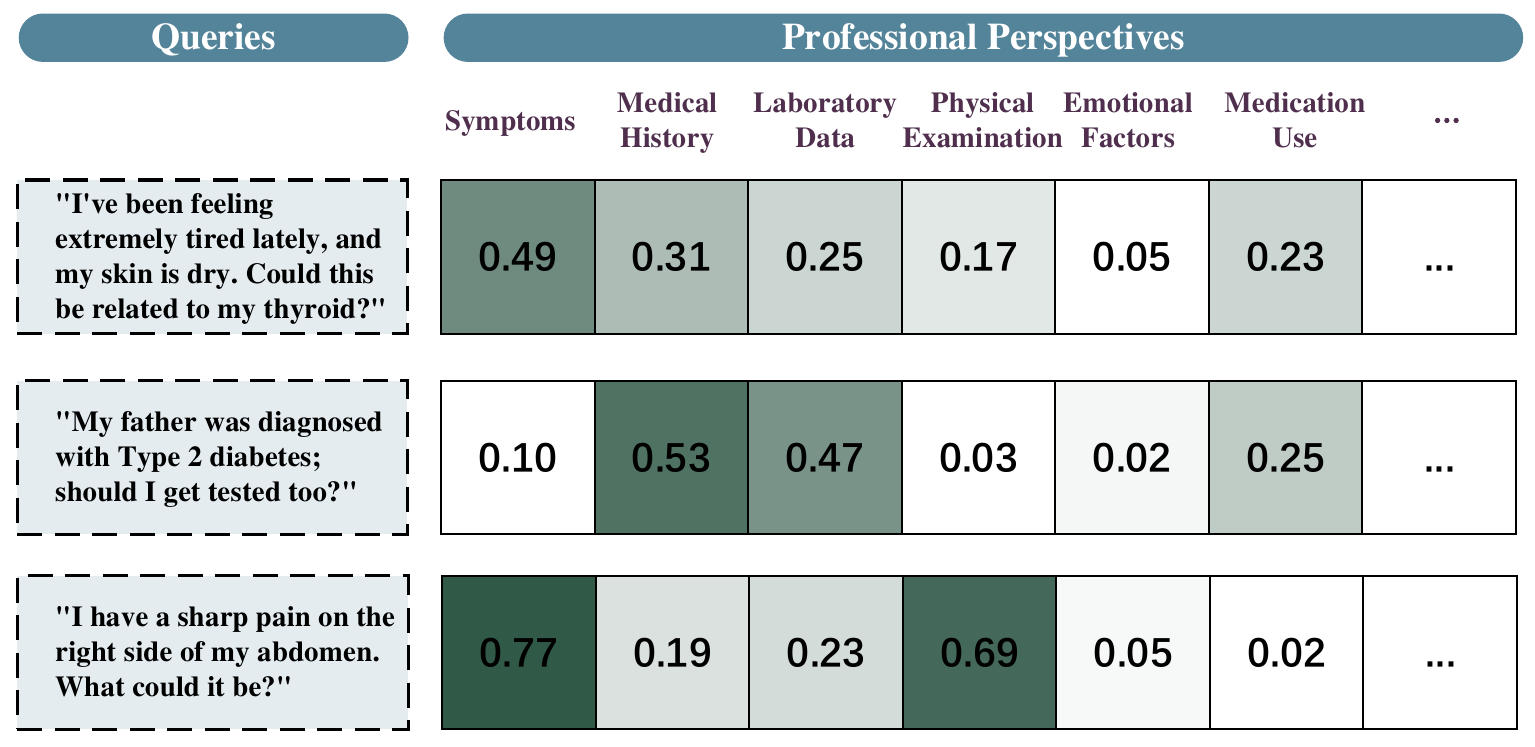} 
  \caption{Visualization of Perspective Vectors, using varying shades of color to represent different normalized scores. The graph displays various scenarios corresponding to different types of queries.}
  \label{fig:matrix}
\end{figure}

\paragraph{Topic Modeling.}

The corpus is transformed into a document-term matrix \( X \in \mathbb{R}^{|C| \times m} \) using Term Frequency-Inverse Document Frequency (TF-IDF), which represents term relevance across documents. Here, \( m \) denotes the total number of unique terms in the corpus, representing the dimensions of the term space. Non-negative Matrix Factorization (NMF) is then applied to \( X \), yielding two non-negative matrices:

\begin{equation}
X \approx W H,
\end{equation}
where \( W \in \mathbb{R}^{|C| \times r} \) encodes document-topic relationships, and \( H \in \mathbb{R}^{r \times m} \) encodes topic-term distributions. The number of topics \( r \) is chosen by maximizing coherence scores on a validation dataset, ensuring optimal interpretability. For each topic \( j \), the top \( k \) terms from the corresponding row \( \mathbf{h}_j \) of \( H \) are extracted:
\begin{equation}
S = \bigcup_{j=1}^{r} \text{Top}_k(\mathbf{h}_j),
\end{equation}
where \( S \) is the set of candidate terms representing each topic. The number of terms \( k \) is fixed at 10 to provide a consistent representation of topics.

Notably, topic terms can be individual words or multi-word phrases, determined by the tokenization method in TF-IDF. In practice, we extract both unigrams and bigrams, which provide richer semantic information and help mitigate the ambiguity of polysemous words. 

\paragraph{Alignment of Perspectives.}

To align extracted topics with the structural components, cosine similarity is computed between each principal component \( \mathbf{v}_k \) and the topic term \( s_j\) in the obtained set \( S \) from NMF. The topic terms  are embedded using the same embedding model to ensure alignment with the dimensionality of \( \mathbf{v}_i \). Each principal component \( \mathbf{v}_i \) is assigned the topic terms \( \mathbf{s}_j \) that maximizes the similarity:
\begin{equation}
\ p_i = \arg \max_{\mathbf{s}_j} \frac{\mathbf{v}_i \cdot \mathbf{s}_j}{\|\mathbf{v}_i\| \|\mathbf{s}_j\|},
\end{equation}
where \( p_i \) represents the professional perspective corresponding to \( \mathbf{v}_i \) and \( \mathbf{s}_j \) represents the embedded vector of the topic term \( s_j \in S \).

As a result of this process, we obtain a set of professional perspectives \( P = \{p_1, p_2, \dots, p_n\} \), where each \( \ p_i \) encapsulates a distinct combination of structural and semantic patterns in the corpus.

\subsection{Intention Recognition and Query Rewriting}\label{sec:rewrite}


After completing the structural analysis and feature extraction of the corpus, we obtain a set of professional perspectives \( P = \{p_1, p_2, \dots, p_n\} \), where each \( p_i \) represents a distinct professional perspective. For a given query \( q \), we compute its similarity to each perspective vector \( p_i \), yielding a weight \( w_i \) that quantifies the alignment between \( q \) and \( p_i \). To exclude weak alignments, weights are filtered by a threshold \( \theta \), which is determined during the offline stage by analyzing the statistical distribution of weights to ensure a consistent balance between relevance and coverage.

The weight \( w_i \) is computed as follows:
\[
w_i = 
\begin{cases} 
  \text{Similarity}(q,  p_i) & \text{if } \text{Similarity}(q,  p_i) > \theta \\
  0 & \text{otherwise},
\end{cases}
\]
where \( \text{Similarity}(q, p_i) \) represents the cosine similarity between the query \( q \) and perspective \( p_i \). 
The threshold \( \theta \) is chosen to balance the inclusion of relevant perspectives with the exclusion of noisy matches. 
In practice, we set \( \theta \) to the 30\%-percentile of the similarity score distribution on a randomly held-out validation set, 
which serves as a heuristic to balance precision and coverage without overfitting to a specific domain.

The resulting weights form a \textit{Perspective Vector} \( V_q \), represented as:
\[
V_q = \begin{bmatrix} w_1 & w_2 & \dots & w_n \end{bmatrix},
\]
where the \( i \)-th element \( w_i \) reflects the relevance of perspective \( p_i \) to the query \( q \).
An illustration of the normalized weights across different query scenarios is shown in Figure~\ref{fig:matrix}.

\textbf{Query Rewriting:} Using the Perspective Vector \( V_q \), the system rewrites the query \( q \) for each perspective \( p_i \) with a non-zero weight \( w_i \). A large-scale language model, referred to as the \textbf{rewriter}, generates rewritten content \( C_i \) for each such perspective:
\begin{equation}
C_i = \text{Rewriter}(q, p_i) \quad \forall \, p_i \, \text{with} \, w_i \neq 0,
\end{equation}
where \( q \) is the original query, \( p_i \) is the specific perspective, and \( w_i \) represents its weight. The rewriting process leverages the contextual understanding of the rewriter model to produce content tailored to the unique nuances of each perspective. To enable scalability across domains, we employ a zero-shot in-context learning strategy, where the rewriter follows a structured prompt template, as shown below:

\begin{tcolorbox}[colback=white, colframe=black, title=Prompt Template]
You are a domain expert specializing in analyzing and rewriting queries to provide comprehensive information from different perspectives. 
Given the following input query and a list of perspectives, generate a pseudo-answer document with one section for each perspective.

\textbf{Input Query:} \{Query\}

\textbf{List of Perspectives:} \{Perspective List\}

\textbf{Generated Pseudo-Answer Document:}  \\
1. \{Perspective 1\}:  \\
2. \{Perspective 2\}:  \\
3. \{Perspective 3\}:  
...
\end{tcolorbox}

This template allows the framework to adapt seamlessly across domains such as law, medicine, and science while maintaining consistent multi-view query generation..

\textbf{Contextual Document Retrieval:} For each rewritten query \( C_i \), the system retrieves a set of documents from the corpus using similarity-based search. The retrieved set \( R_i \) for perspective \( p_i \) is defined as:
\begin{equation}
  R_i = \left\{ d_{ij} \,\middle|\, 
  \begin{aligned}
    & d_{ij} \text{ ranks among the top } k \text{ by } \\
    & \text{Similarity}(d_{ij}, C_i), \forall j \in \{1, \ldots, k\}
  \end{aligned}
  \right\},
\end{equation}
where \( d_{ij} \) represents the \( j \)-th document in \( R_i \), and \( \text{Similarity}(d_{ij}, C_i) \) is the cosine similarity between document \( d_{ij} \) and rewritten query \( C_i \). \( k \) is a hyperparameter that defines the retrieval depth, dynamically set based on the user's requirements for the desired scope and granularity of the retrieved results.

The retrieval process produces a collection of document sets \( \{R_1, R_2, \dots, R_n\} \), each tailored to a specific professional perspective. This ensures the query is contextually expanded and aligned with the multifaceted aspects of the domain, significantly enhancing the relevance and specificity of the retrieved information.

\subsection{Retrieval Augmentation and Final Inference}

After assembling the retrieved document sets \(\{R_1, R_2, \ldots, R_n\}\), the system refines these results to enhance relevance and utility. This process is guided by the original \textit{Perspective Vector} \( V_q \), ensuring that the final output reflects the multi-perspective nature of the query.

\textbf{Re-ranking Documents:} The system recalculates the relevance of each document \( d_{ij} \) in the retrieved sets based on its alignment with the rewritten query \( C_i \) and the corresponding perspective weight \( w_i \). The relevance score \( \text{Rel}(d_{ij}) \) is defined as:
\begin{equation}
  \text{Rel}(d_{ij}) = \frac{\text{Similarity}(d_{ij}, C_i)}{w_i},
\end{equation}
where \( \text{Similarity}(d_{ij}, C_i) \) measures the cosine similarity between the document \( d_{ij} \) and the rewritten query \( C_i \), and \( w_i \) represents the weight of perspective \( p_i \) in \( V_q \). This formula ensures that documents strongly aligned with both the rewritten query and the perspective are prioritized.

\textbf{Structured Prompt Generation:} The re-ranked documents are integrated into a structured prompt \( \mathcal{P} \), combining the original query \( q \) with the restructured document content:
\begin{equation}
  \mathcal{P} = q \, \bowtie \, \bigoplus_{i=1}^{n} \bigoplus_{j=1}^{k_i} d_{ij}^{\prime},
\end{equation}
where \( d_{ij}^{\prime} \) are the re-ranked documents, \( \bowtie \) represents concatenation, and \( \bigoplus \) denotes the combination of results across all perspectives. The prompt ensures that all relevant perspectives are represented, with emphasis on their respective importance as determined by \( V_q \).

\textbf{Final Inference:} The structured prompt \( \mathcal{P} \) is fed into a large-scale reader model \( \mathcal{M} \), such as a pretrained transformer-based language model, to generate the final response. The reader model processes the combined information to produce a nuanced, multi-perspective answer that is contextually aligned with the original query’s complexities.

This retrieval augmentation framework provides a comprehensive analysis of the query, combining diverse perspectives while ensuring specificity and contextual richness. The result is a tailored response that effectively addresses the complexities of specialized domains.

\begin{table*}[ht]
  \centering
  \caption{Performance of retrieval methods across datasets. \\Metrics include recall (R) at 5 and 10 documents and mean reciprocal rank (MRR) at 10.}
  \small
  \renewcommand{\arraystretch}{1.1}
  \setlength{\tabcolsep}{2pt}
  \begin{tabular}{lcccccccccccc}  
  \hline
  & \multicolumn{3}{c}{\textbf{E-commerce}} & \multicolumn{3}{c}{\textbf{Medical}} 
  & \multicolumn{3}{c}{\textbf{Entertainment}} & \multicolumn{3}{c}{\textbf{Legal}} \\
  \textbf{Method} & R@5 & R@10 & MRR@10 & R@5 & R@10 & MRR@10 & R@5 & R@10 & MRR@10 & R@5 & R@10 & MRR@10 \\
  \hline
  \multicolumn{13}{c}{\textbf{Traditional Models}} \\
  \hline
  BM25 & 20.50\% & 26.00\% & 0.1575 & 31.60\% & 35.20\% & 0.2695 & 31.50\% & 40.10\% & 0.2198 & 15.12\% & 24.81\% & 0.1473 \\
  QLD & 21.70\% & 24.70\% & 0.1629 & 36.10\% & 42.20\% & 0.2385 & 25.20\% & 26.10\% & 0.2032 & 14.46\% & 25.38\% & 0.1389 \\
  \hline
  \multicolumn{13}{c}{\textbf{Dense Retrieval Models}} \\
  \hline
  Bert & 2.90\% & 4.30\% & 0.0257 & 1.60\% & 2.30\% & 0.0167 & 4.70\% & 6.00\% & 0.0403 & 4.37\% & 6.32\% & 0.0592 \\
  BGE-Large & 45.60\% & 54.40\% & \textbf{0.3518} & 55.10\% & 59.60\% & 0.4885 & 43.10\% & \textbf{53.80\%} & \textbf{0.3106} & 11.10\% & 18.30\% & 0.0913 \\
  BGE-M3 & 40.30\% & 48.30\% & 0.2977 & 53.60\% & 57.50\% & 0.4694 & 31.80\% & 42.10\% & 0.2366 & 12.09\% & 20.63\% & 0.1153 \\
  \hline
  \multicolumn{13}{c}{\textbf{Dense Retrieval Models with MVRAG}} \\
  \hline
  Bert & 2.90\% & 5.80\% & 0.0307 & 9.00\% & 11.70\% & 0.0352 & 4.70\% & 7.20\% & 0.0390 & 2.23\% & 8.34\% & 0.0351 \\
  BGE-Large & \textbf{48.80\%} & \textbf{66.80\%} & 0.3314 & 51.60\% & 59.40\% & 0.4550 & \textbf{45.90\%} & 48.40\% & 0.2183 & 13.30\% & 24.95\% & 0.1087 \\
  BGE-M3 & 40.10\% & 56.10\% & 0.2910 & \textbf{60.40\%} & \textbf{69.40\%} & \textbf{0.5123} & 37.40\% & 42.40\% & 0.2413 & \textbf{16.09\%} & \textbf{29.18\%} & \textbf{0.1525} \\
  \hline
  \end{tabular}
  
  \label{tab:results_retrieval}
\end{table*}

\section{Experiments}

\subsection{Retrieval Task}

\subsubsection{Dataset and Experiment Setup}

For the retrieval task, the MVRAG framework aims to retrieve diverse and comprehensive documents to better support downstream generation tasks, differing from traditional methods that focus on precision. Despite this distinction, experimental results show that MVRAG performs on par with or even surpasses traditional retrieval methods.

The evaluation was conducted using datasets from four domains: \textbf{e-commerce}, \textbf{medical}, \textbf{entertainment video}, and \textbf{legal}. The first three domains were sourced from Alibaba’s \textit{Multi-CPR}\cite{long2022multi}, containing approximately one million records and 1000 queries per domain. For the legal domain, the \textit{LeCaRDv2} dataset~\cite{li2023lecardv2}, a benchmark dataset comprising 800 queries and 55,192 criminal case document candidates, was used.

To ensure a fair comparison, the baselines included traditional retrieval models BM25 and QLD, implemented with \textit{Pyserini}\footnote{\url{https://github.com/castorini/pyserini}} default settings, and dense retrieval models such as \textbf{bert-base-chinese}\footnote{\url{https://huggingface.co/google-bert/bert-base-chinese}}, \textbf{bge-large-zh-v1.5}\footnote{\url{http://huggingface.co/BAAI/bge-large-zh-v1.5}}, and \textbf{bge-m3}\footnote{\url{https://huggingface.co/BAAI/bge-m3}}, configured according to standard \textit{Dense} framework practices. In MVRAG, the same dense retrieval models and \textit{Dense}\footnote{\url{https://github.com/luyug/Dense}} framework were employed alongside \textbf{DeepSeek}\footnote{\url{https://www.deepseek.com/}} as the query rewriter, with the retrieval depth  k  for each perspective set to 10.

\subsubsection{Results}

The experimental results demonstrate that the MVRAG framework, designed to retrieve diverse and comprehensive documents to support downstream tasks, achieves competitive performance across various datasets. While minor declines are observed in certain metrics, the majority remain comparable to or surpass baseline methods, particularly in knowledge-intensive domains.

In the \textbf{medical domain}, Recall@10 improves significantly from 57.50\% to 69.40\%, with MRR@10 increasing from 0.4694 to 0.5123. Similarly, in the \textbf{legal domain}, Recall@10 increases from 20.63\% to 29.18\%, and MRR@10 improves from 0.1153 to 0.1525. In less complex domains such as \textbf{e-commerce} and \textbf{entertainment video}, MVRAG consistently maintains or slightly improves performance, achieving a Recall@10 of 66.80\% in e-commerce.

These findings underscore MVRAG’s robust retrieval capabilities across a wide range of datasets, even though its primary objective is to enhance the diversity and comprehensiveness of retrieved documents in knowledge-intensive scenarios.
\begin{table*}[ht]
  \centering
  \small
  \caption{Performance on legal and medical generation tasks.\\ The abbreviations are as follows: MS (Medicine Single), LS (Legal Single), LM (Exact) (Exact match accuracy for Legal Multi-choice), LM (Percentage) (Percentage of correct answers for Legal Multi-choice), and LSub (Legal Subjective QA).}
  \begin{tabular}{llcccccc}
  \label{table:generation}
  \textbf{Model} & \textbf{Method} & \textbf{MS} & \textbf{LS} & \textbf{LM (Exact)} & \textbf{LM (Percentage)} & \textbf{LSub} \\
  \hline
  \multirow{6}{*}{\textbf{GLM-4-9B}} 
  & Baseline & 68.72\% & 66.42\% & 5.49\% & 14.57\% & 18.5\% \\
  & RAG & 74.39\% & 69.41\% & 7.03\% & 18.43\% & 19.35\% \\
  & Query2doc & 78.31\% & \textbf{73.10\%} & 13.93\% & 31.33\% & 22.94\% \\
  & MultiQuery  & 75.02\% & 70.43\% & 7.00\% & 23.58\% & 21.38\% \\
  & FLARE & \textbf{80.37\%} & 67.23\% & \underline{15.59\%} & \underline{37.94\%} & \underline{28.35\%} \\
  & MVRAG & \underline{80.20\%} & \underline{72.50\%} & \textbf{22.91\%} & \textbf{48.42\%} & \textbf{29.64\%} \\
  \hline
  \multirow{6}{*}{\textbf{Qwen-7B}} 
  & Baseline & 66.50\% & 72.13\% & 5.00\% & 14.57\% & 14.91\% \\
  & RAG & 70.21\% & 75.76\% & 5.56\% & 18.89\% & 16.12\% \\
  & Query2doc & \textbf{79.01\%} & \underline{76.94\%} & 12.30\% & 35.12\% & 19.35\% \\
  & MultiQuery  & 77.96\% & 77.02\% & 13.14\% & 35.19\% & 20.72\% \\
  & FLARE  & 78.61\% & 76.56\% & \underline{24.57\%} & \underline{45.03\%} & \underline{22.21\%} \\
  & MVRAG & \underline{78.83\%} & \textbf{77.89\%} & \textbf{26.26\%} & \textbf{50.59\%} & \textbf{30.69\%} \\
  \hline
  \end{tabular}
  
  \label{table:merged_vertical_results}
\end{table*}

\subsection{Generation Task}

\subsubsection{Dataset and Experiment Setup}
For the generation tasks, we demonstrated the improvements from the MVRAG framework using domain-specific question-answering tasks. In the legal domain, we used the \textit{JEC-QA} dataset~\cite{zhong2020jec}, which contains 26,365 multiple-choice questions from the Chinese judicial examination, a challenging test for legal practitioners. The dataset was split into single-choice and multiple-choice questions for testing with LLMs, using the provided legal texts as the retrieval database. In the medical domain, we selected the \textit{MED-QA} dataset\footnote{\url{https://huggingface.co/datasets/bigbio/med_qa}}, consisting of 14,123 single-choice biomedical questions from the National Medical Licensing Examination.

We employed GLM-4-9B\footnote{\url{https://github.com/THUDM/GLM-4}.} and Qwen-7B\footnote{\url{https://modelscope.cn/models/qwen/Qwen-7B}.} as rewriter and generator models for their strong performance in Chinese and efficient operation. Evaluation was based on accuracy for single-choice questions, and exact match accuracy and correct answer percentage for multiple-choice questions. For subjective questions, we used official standard answers and an automatic scoring system based on keyword matching. The retrieval depth  for other models was set to 8, while for MVRAG, the retrieval depth k for each perspective was set to 4 and after re-ranking the final retrieval depth was set to 8. Our experiments were conducted in an environment equipped with a single NVIDIA A100-PCIE-40GB GPU.
\subsubsection{Results} 
The results, shown in Table~\ref{table:merged_vertical_results}, demonstrate that MVRAG excels across tasks, particularly in complex scenarios. While single-choice questions showed relatively smaller gains due to well-defined answers, MVRAG still improved accuracy to 72.50\% on GLM-4-9B and 77.89\% on Qwen-7B. However, the most notable improvements were seen in multiple-choice and subjective tasks, where MVRAG’s multi-view query rewriting and feature extraction significantly enhanced model performance. For legal multiple-choice tasks, MVRAG boosted accuracy to 22.91\% on GLM-4-9B and 26.26\% on Qwen-7B, nearly tripling the baseline performance. 

We also observe that FLARE, with its dense retrieval mechanism, performs strongly on relatively straightforward tasks requiring precise factual knowledge, such as medical single-choice (MS). 
However, in more complex scenarios such as legal multiple-choice and subjective questions, FLARE falls short of MVRAG. 
This is because MVRAG contributes not only additional knowledge but also multi-dimensional perspectives, which provide richer context and more nuanced reasoning support, particularly beneficial in subjective tasks where deeper inference is required. 

MVRAG’s strength lies in its ability to integrate domain-specific perspectives, enabling deeper analysis of complex legal and medical scenarios. By capturing multidimensional aspects like dispute focus in legal cases or medical history in diagnoses, MVRAG delivers more precise and contextually rich results. This multi-view approach explains why MVRAG consistently outperforms other methods in tasks requiring complex reasoning.

In summary, MVRAG demonstrates value in simpler tasks but shows transformative improvements in complex, knowledge-intensive queries, enhancing retrieval precision and inference quality in domains like law and medicine.

\subsection{Efficiency Analysis}
\begin{table}[!t]
\centering
\small
\caption{Latency breakdown of MVRAG components. Measurements are on a NVIDIA A100-PCIE-40GB GPU.}
\begin{tabular}{llc}
\hline
\textbf{Phase} & \textbf{Component} & \textbf{Time} \\ \hline
\multirow{2}{*}{Offline} 
  & Perspective extraction (ML)  & 1.5 min \\
  & Perspective extraction (LLM) & 2 h \\ \hline
\multirow{5}{*}{Online} 
  & Perspective-based rewriting  & 0.90 s \\
  & Multi-view retrieval         & 0.32 s \\
  & Re-ranking                   & 0.25 s \\
  & Reader (answer generation)   & 1.10 s \\
  & Pre/post-processing          & 0.30 s \\ 
\cline{2-3}
  & \textbf{Total (online)}      & \textbf{2.87 s} \\ \hline
\end{tabular}
\label{tab:latency_breakdown_multirow}
\end{table}

We evaluate the computational characteristics of MVRAG using GLM-4-9B on a single A100 GPU, reporting both offline preprocessing costs and online inference latency in Table~\ref{tab:latency_breakdown_multirow}.

For the offline stage, perspective extraction is performed only once per corpus. 
The ML-based method completes this step in about 1.5 minutes and produces perspectives such as \textit{Basic Fact}, \textit{Focus of Dispute}, \textit{Application of Law}, \textit{Penalty}, and \textit{Criminal History}, 
leading to downstream performance of 48.42\% on Legal Multiple-choice and 29.64\% on Legal Subjective QA. 
By contrast, the LLM-based method requires about 2 hours and yields perspectives like \textit{Essential Case Facts}, \textit{Primary Legal Issue}, \textit{Legal Precedent Interpretation}, \textit{Criminal History}, and \textit{Sanction}, 
but achieves lower accuracy (39.98\% on Legal Multiple-choice and 26.45\% on Legal Subjective QA). 
This comparison highlights that while the LLM-based variant can generate more nuanced perspectives, 
the ML-based approach offers a more favorable trade-off between efficiency and effectiveness and is thus used as the default setting.  

During online inference, perspective-based rewriting adds 0.9s, retrieval 0.32s, and re-ranking 0.25s, while the reader dominates runtime with about 1.1s per query. 
The total online latency is under 3s in our experimental setup, indicating that the framework is practical for offline and semi-real-time settings, with manageable additional cost compared to vanilla RAG. 
For comparison, the FLARE method achieves relatively close performance to MVRAG but requires multiple iterative retrieval–generation cycles, which results in a noticeably higher overall runtime.
\begin{table}[t]
\centering
\caption{Effect of using different LLMs as the \textit{query rewriter}. 
Results are reported on Medical single-choice (MS) and Legal subjective (LSub)}
\label{tab:rewriter-ablation}
\begin{tabular}{lcc}
\toprule
Rewriter (LLM) & MS (Acc.\%) & LSub (Score) \\
\midrule
DeepSeek    & \textbf{80.20} & 29.64 \\
GPT-4       & 79.35 & \textbf{30.12} \\
GLM-4-9B    & 79.82 & 29.41 \\
Qwen-7B     & 72.87 & 26.95 \\
\bottomrule
\end{tabular}
\end{table}
\subsection{Ablation Experiments}
\begin{figure*}[htbp]
\centering
\includegraphics[width=0.8\textwidth,  page=1]{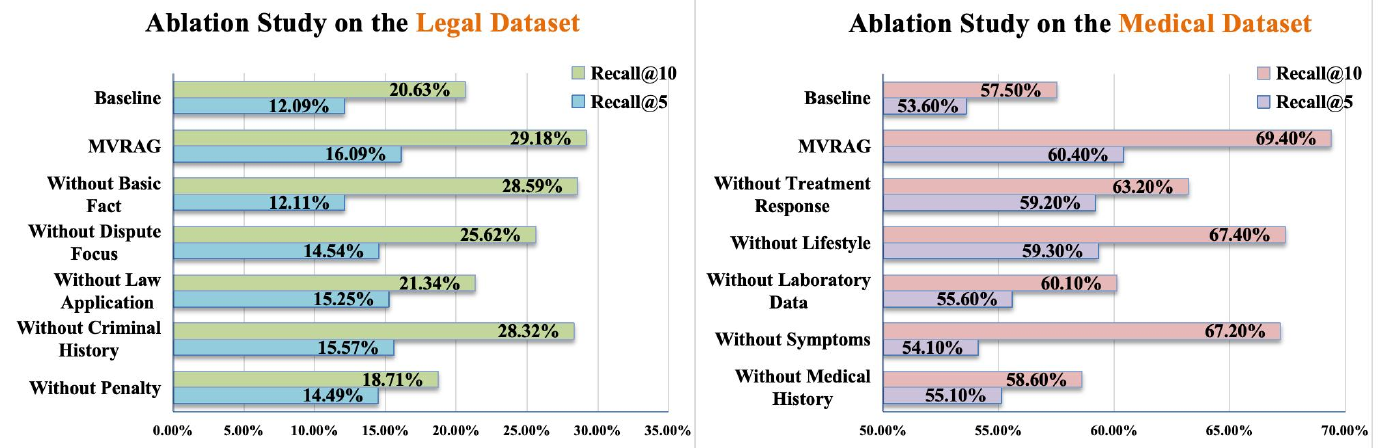}
\caption{Ablation Study on the impact of perspective selection strategies in our framework on Medical and Legal datasets. In the legal domain, the chart shows Recall@5 and Recall@10 after excluding each perspective: \textit{Basic Fact, Focus of Dispute, Application of Law, Penalty, Criminal History}. For the medical domain, it displays the effects of removing \textit{Medical History, Symptoms, Laboratory Data, Treatment Response, Lifestyle}. Each bar indicates the performance impact versus the full baseline and direct retrieval.}
\label{fig:ablation}
\end{figure*}

We conducted ablation experiments to better understand the contribution of individual components in our framework. 
This analysis focuses on two aspects: the effect of different perspectives and the role of the query rewriter. 

\paragraph{Effect of Different Perspectives.}
Figure~\ref{fig:ablation} presents results on the medical and legal domains when excluding individual perspectives. 
In the medical domain, removing \textit{Medical History} led to the largest performance drop, while \textit{Lifestyle} had the least impact. 
In the legal domain, \textit{Fundamental Facts} proved most critical, whereas \textit{Penalty} contributed the least. 
These findings demonstrate that different perspectives carry varying importance depending on the domain, which is consistent with our design of dynamically weighting perspectives according to their relevance. 

\paragraph{Effect of Different Rewriters.}
We further ablated the query rewriter while fixing \textit{GLM-4-9B} as the reader model. 
Table~\ref{tab:rewriter-ablation} reports results when varying the LLM used for rewriting. 
DeepSeek performs best on the objective medical single-choice task (MS), while GPT-4 shows an advantage on the subjective legal task (LSub), with GLM-4-9B competitive in both cases due to its alignment with the reader model.
These results indicate that the rewriter primarily serves to enhance query–perspective alignment, with different LLMs exhibiting complementary strengths across task types, rather than acting as the main source of performance gains.

\subsection{Generalizability Analysis}
\begin{table}[!t]
\centering
\small
\caption{Performance of retrieval methods on the PMC-Patients English dataset.}
\label{tab:pmc}
\begin{tabular}{c l c c c}
\toprule
\textbf{Model} & \textbf{Method} & \textbf{R@5} & \textbf{R@10} & \textbf{MRR@10} \\
\midrule
\multirow{3}{*}{gte-large}
  & Baseline   & 6.16 & 9.79  & 0.06 \\
  & MultiQuery & 9.36 & 13.95 & 0.09 \\
  & \textbf{MVRAG} & \textbf{12.05} & \textbf{17.30} & \textbf{0.12} \\
\midrule
\multirow{3}{*}{bge-large-en}
  & Baseline   & 8.40 & 12.71 & 0.09 \\
  & MultiQuery & 7.78 & 11.95 & 0.08 \\
  & \textbf{MVRAG} & \textbf{14.15} & \textbf{19.35} & \textbf{0.13} \\
\midrule
\multirow{3}{*}{all-MiniLM-L6}
  & Baseline   & 5.46 & 8.94  & 0.05 \\
  & MultiQuery & 6.82 & 10.78 & 0.07 \\
  & \textbf{MVRAG} & \textbf{7.13} & \textbf{11.05} & \textbf{0.08} \\
\bottomrule
\end{tabular}
\end{table}
To further evaluate the generalizability of MVRAG across languages and domains, we conduct experiments on the PMC-Patients dataset~\cite{zhao2022pmc}, which contains 167k patient summaries and 293k patient-patient similarity annotations defined by the PubMed citation graph, serving as a large-scale English medical case retrieval benchmark. 
We employ three representative English embedding models: 
\texttt{gte-large}\footnote{\url{https://huggingface.co/thenlper/gte-large}}, 
\texttt{bge-large-en}\footnote{\url{https://huggingface.co/BAAI/bge-large-en}}, 
and \texttt{all-MiniLM-L6}\footnote{\url{https://huggingface.co/sentence-transformers/all-MiniLM-L6-v2}}. 
As shown in Table~\ref{tab:pmc}, MVRAG consistently achieves substantial 
improvements over both the baselines and MultiQuery across Recall@5, Recall@10, 
and MRR@10 metrics, highlighting its robust \textbf{cross-lingual generalizability}. 
In addition to the core domains evaluated in the main experiments, 
we further conducted case studies in more fields like physics, finance, and geography. 
Representative examples are reported in Section~\ref{app:medical case}, 
demonstrating that our framework can be seamlessly extended to diverse 
knowledge-intensive fields.

\subsection{Fine-Grained Analysis}
\begin{figure*}[htbp]
  \centering
  \includegraphics[width=0.6\textwidth]{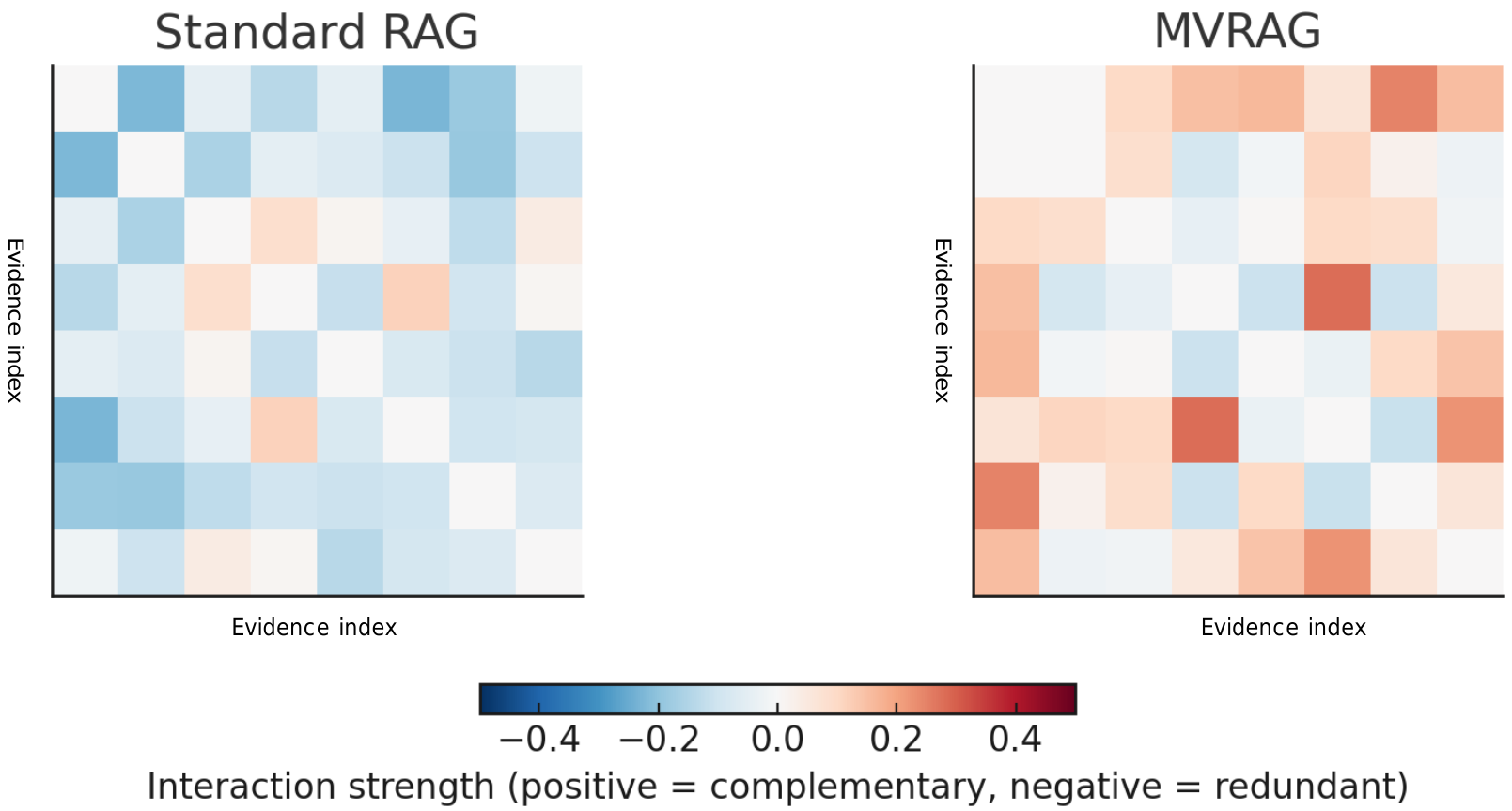}
  \caption{nteraction heatmaps comparing standard RAG and MVRAG. 
  Each cell encodes the pairwise interaction strength between two retrieved evidences, estimated using a double leave-one-out strategy. 
  Blue cells indicate negative interactions (redundancy, i.e., evidences providing overlapping information), 
  while red cells indicate positive interactions (complementarity, i.e., evidences providing distinct and reinforcing perspectives). }
  \label{fig:redundant}
\end{figure*}
To further examine the principles behind MVRAG beyond aggregate metrics and case studies, we construct fine-grained interaction heatmaps that visualize how retrieved evidences affect each other. 
For subjective questions, we compute per-instance scores with and without specific evidences, and estimate pairwise interaction strength using a double leave-one-out strategy. This procedure allows us to quantify whether two evidences are complementary (positive interaction) or redundant (negative interaction).

Figure~\ref{fig:redundant} contrasts the interaction patterns of a standard RAG system and our proposed MVRAG. In standard RAG, many evidences are retrieved from semantically similar passages, resulting in substantial redundancy (large negative interactions). In contrast, MVRAG yields a more balanced structure: within-perspective evidences still exhibit minor redundancy, but cross-perspective evidences are more often complementary, yielding positive interactions. This demonstrates that perspective-guided rewriting and re-ranking do not merely add more evidences, but actively diversify the retrieval space and enhance cross-perspective synergy. Therefore, MVRAG not only improves coverage but also reshapes the interaction structure of evidences in a way that 
better supports reasoning.
\section{Case Study}
\label{app:medical case}

Within this section, we present case studies from multiple domains. For the medical domain, we employ the \textit{PMC-Patients} dataset as the retrieval database. As illustrated in Figure~\ref{fig:casestudy}, we selected a case of \textbf{Huntington's disease} from \textit{PubMed} and transformed it into a query voiced by the patient, serving as our original query. In scenarios absent of our multi-view retrieval model, the retrieval system primarily fetched articles related to superficial symptoms such as emotional dysregulation and movement disorders. This surface-level matching, due to similar characterizations, erroneously led to articles on \textbf{Vitamin B12 deficiency}, consequently misdiagnosing the condition as a \textbf{Vitamin B12 deficiency}.

Conversely, our multi-view retrieval system initiates with intent recognition, prioritizing symptoms and medical history for query reformulation. The perspectives utilized by the framework are determined during an offline stage, where machine learning techniques such as Principal Component Analysis (PCA) and Non-negative Matrix Factorization (NMF) analyze the structural and semantic patterns of the domain-specific corpus. This automated process extracts professional perspectives, ensuring they are representative of the corpus's inherent knowledge while being adaptable to various domains.

Through individual matching and re-ranking, the system yielded comparatively relevant search outcomes. Specifically, symptom keywords, refined with greater precision, directed the search towards articles on neurodegenerative diseases. Simultaneously, the emphasis on similar family medical history surfaced articles on genetic testing. With the aid of these references, the model adeptly identified the potential for \textbf{Huntington's disease}.

This case study emphatically demonstrates the superiority of our multi-view retrieval model. By leveraging perspectives refined through offline corpus analysis and multi-view reformulation, the system adeptly navigated towards more pertinent and informative references, thereby significantly enhancing the accuracy of the final diagnostic inference.

Importantly, this effect is not confined to the medical domain. 
For instance, Table~\ref{tab:finance_case} illustrates a financial case study on Basel~III, 
where a standard RAG system, focusing narrowly on capital requirement documents, 
incorrectly suggested that bank capital adequacy is reliably ensured. 
In contrast, MVRAG incorporated perspectives such as \emph{Regulatory Rules}, 
\emph{Risk Categories}, and \emph{Macro Conditions}, retrieving complementary evidence 
on hybrid instruments and systemic risk. 
This broader evidence base enabled MVRAG to provide a more nuanced and accurate conclusion, 
demonstrating that the benefits of perspective-guided retrieval extend beyond medicine 
to other knowledge-intensive fields such as finance.
\begin{table}[t]
\centering
\scriptsize
\caption{\centering A financial case study.}
\begin{tabular}{p{0.16\linewidth} | p{0.32\linewidth} | p{0.32\linewidth}}
\toprule
\textbf{Component} & \textbf{Standard RAG} & \textbf{MVRAG } \\
\midrule
\textbf{Query} & \multicolumn{2}{c}{\parbox{0.65\linewidth}{\centering ``What are the main factors that challenge the reliability of bank capital adequacy assessments under Basel III?''}} \\
\midrule
\textbf{Professional Perspectives} 
& Not considered. 
& \{Regulatory Rules, Risk Categories, Macro Conditions, Financial Instruments\} \\
\midrule
\textbf{Retrieved Evidences} 
& \parbox{\linewidth}{
-- Basel III guidelines on minimum capital requirements.\\
-- Supervisory reports on capital buffers.\\
-- Studies highlighting Tier 1 capital definitions.\\
-- Commentary on the leverage ratio and risk weights.
} 
& \parbox{\linewidth}{
-- Basel III rules on AT1 instruments.\\
-- Studies on CoCo bond trigger mechanisms.\\
-- Evidence on capital procyclicality.\\
-- Macro-level analyses of systemic risk exposure.
} \\
\midrule
\textbf{Answer} 
& \parbox{\linewidth}{
``Basel III enhances reliability mainly through \textbf{stricter capital rules}.  
Higher minimum ratios, the introduction of capital conservation buffers, and clearer Tier 1 definitions collectively strengthen the framework.  
Although effective in improving stability, reliability is still portrayed largely as a function of \textbf{stronger capital requirements}.''  
} 
& \parbox{\linewidth}{
``Reliability is challenged not only by capital definitions but also by broader \textbf{systemic dynamics}.  
Conditional \textbf{loss-absorption} of hybrid instruments introduces uncertainty, while exposure to \textbf{systemic risk} and the \textbf{procyclical} behavior of requirements under downturns further undermine stability.  
This highlights that reliability depends on both regulatory design and external macro-financial conditions.''  
} \\
\bottomrule
\end{tabular}
\label{tab:finance_case}
\end{table}

\section{Conclusion}

\noindent In this paper, we propose a multi-perspective approach to Retrieval-Augmented Generation tailored for knowledge-dense domains, aiming to incorporate domain-specific insights missing from existing methods and enhance the reliability and interpretability of retrieval outcomes. By employing professional perspectives extraction, intent recognition, query rewriting, and document re-ranking, We have significantly improved generation performance while maintaining retrieval accuracy in fields such as law and medicine. Through experiments, we demonstrate the impact of integrating multi-perspective information, laying the groundwork for future incorporation of machine learning techniques into RAG systems. Our approach introduces advanced feature extraction techniques and unleashes the potential of multi-view information, accelerating the application of LLMs in knowledge-dense fields.

\section*{Acknowledgement}

\noindent This work was supported by the National Science Fund for Excellent Young Scholars (Overseas) under grant No.\ KZ37117501, National Natural Science Foundation of China ( No. 62306024), and Xiaomi’s “Open bidding for selecting the best candidates” project.



\begin{thebibliography}{1}
\bibliographystyle{IEEEtran}

\bibitem{gao2023retrieval}
Y. Gao, Y. Xiong, X. Gao, K. Jia, J. Pan, Y. Bi, Y. Dai, J. Sun, and H. Wang, ‘‘Retrieval-augmented generation for large language models: A survey,’’ \textit{arXiv preprint arXiv:2312.10997}, 2023.

\bibitem{wang2023query2doc}
L. Wang, N. Yang, and F. Wei, ‘‘Query2doc: Query expansion with large language models,’’ \textit{arXiv preprint arXiv:2303.07678}, 2023.

\bibitem{ma2023query}
X. Ma, Y. Gong, P. He, H. Zhao, and N. Duan, ‘‘Query rewriting for retrieval-augmented large language models,’’ \textit{arXiv preprint arXiv:2305.14283}, 2023.

\bibitem{langchain2024multiquery}
LangChain, ``MultiQueryRetriever documentation,'' \textit{Online}, available: \url{https://python.langchain.com/v0.2/docs/how_to/MultiQueryRetriever}, accessed Jan. 3, 2024.

\bibitem{dong2024mc}
K. Dong, D. G. X. Deik, Y. Q. Lee, H. Zhang, X. Li, C. Zhang, and Y. Liu, ``MC-indexing: Effective long document retrieval via multi-view content-aware indexing,'' in \textit{Findings of the Association for Computational Linguistics: EMNLP 2024}, pp. 2673--2691, 2024.

\bibitem{cao2025neusym}
R. Cao, H. Zhang, T. Huang, Z. Kang, Y. Zhang, L. Sun, H. Li, Y. Miao, S. Fan, L. Chen, and K. Yu, ``NeuSym-RAG: Hybrid neural symbolic retrieval with multiview structuring for PDF question answering,'' \textit{arXiv preprint arXiv:2505.19754}, 2025.

\bibitem{xu2025multigranularity}
Y. Xu, et al.,
“A Multi-Granularity Multimodal Retrieval Framework for Multimodal Document Tasks,” 
arXiv preprint arXiv:2505.01457, 2025.

\bibitem{izacard2022atlas}
G. Izacard, P. Lewis, M. Lomeli, L. Hosseini, F. Petroni, T. Schick, J. Dwivedi-Yu, A. Joulin, S. Riedel, and E. Grave, ‘‘Atlas: Few-shot learning with retrieval augmented language models,’’ \textit{arXiv preprint arXiv:2208.03299}, 2022.

\bibitem{jiang2023flare}
Z. Jiang, F. Xu, X. Li, W. Xiong, J. Callan, and G. Neubig, 
``Active Retrieval Augmented Generation,'' 
\textit{arXiv preprint arXiv:2305.06983}, 2023.

\bibitem{huo2023retrieving}
S. Huo, N. Arabzadeh, and C. LA Clarke, ‘‘Retrieving supporting evidence for LLMs generated answers,’’ \textit{arXiv preprint arXiv:2306.13781}, 2023


\bibitem{guu2020retrieval}
K. Guu, K. Lee, Z. Tung, P. Pasupat, and M. Chang, ‘‘Retrieval augmented language model pre-training,’’ in \textit{International conference on machine learning}, 2020, pp. 3929--3938.

\bibitem{lewis2020retrieval}
P. Lewis, E. Perez, A. Piktus, F. Petroni, V. Karpukhin, N. Goyal, H. Kuttler, M. Lewis, W.-t. Yih, T. Rocktäschel, et al., ‘‘Retrieval-augmented generation for knowledge-intensive NLP tasks,’’ \textit{Advances in Neural Information Processing Systems}, vol. 33, pp. 9459--9474, 2020.
\bibitem{khattab2021relevance}
O. Khattab, C. Potts, and M. Zaharia, ‘‘Relevance-guided supervision for openQA with Colbert,’’ \textit{Transactions of the Association for Computational Linguistics}, vol. 9, pp. 929--944, 2021.

\bibitem{shuster2021retrieval}
K. Shuster, S. Poff, M. Chen, D. Kiela, and J. Weston, ‘‘Retrieval augmentation reduces hallucination in conversation,’’ \textit{arXiv preprint arXiv:2104.07567}, 2021.
\bibitem{saad2023ares}
J. Saad-Falcon, O. Khattab, C. Potts, and M. Zaharia, ‘‘Ares: An automated evaluation framework for retrieval-augmented generation systems,’’ \textit{arXiv preprint arXiv:2311.09476}, 2023.
\bibitem{ram2023context}
O. Ram, Y. Levine, I. Dalmedigos, D. Muhlgay, A. Shashua, K. Leyton-Brown, and Y. Shoham, ‘‘In-context retrieval-augmented language models,’’ \textit{Transactions of the Association for Computational Linguistics}, vol. 11, pp. 1316--1331, 2023.
\bibitem{es2023ragas}
S. Es, J. James, L. Espinosa-Anke, and S. Schockaert, ‘‘Ragas: Automated evaluation of retrieval augmented generation,’’ \textit{arXiv preprint arXiv:2309.15217}, 2023.
\bibitem{zhu2023large}
Y. Zhu, H. Yuan, S. Wang, J. Liu, W. Liu, C. Deng, Z. Dou, and J.-R. Wen, ‘‘Large language models for information retrieval: A survey,’’ \textit{arXiv preprint arXiv:2308.07107}, 2023.

\bibitem{zhao2023survey}
W. Xin Zhao, K. Zhou, J. Li, T. Tang, X. Wang, Y. Hou, Y. Min, B. Zhang, J. Zhang, Z. Dong, et al., ‘‘A survey of large language models,’’ \textit{arXiv preprint arXiv:2303.18223}, 2023.

\bibitem{yao2023knowledge}
J. Yao, W. Xu, J. Lian, X. Wang, X. Yi, and X. Xie, ‘‘Knowledge Plugins: Enhancing Large Language Models for Domain-Specific Recommendations,’’ \textit{arXiv preprint arXiv:2311.10779}, 2023.

\bibitem{xi2023towards}
Y. Xi, W. Liu, J. Lin, J. Zhu, B. Chen, R. Tang, W. Zhang, R. Zhang, and Y. Yu, ‘‘Towards open-world recommendation with knowledge augmentation from large language models,’’ \textit{arXiv preprint arXiv:2306.10933}, 2023.

\bibitem{chen2023benchmarking}
J. Chen, H. Lin, X. Han, and L. Sun, ‘‘Benchmarking large language models in retrieval-augmented generation,’’ \textit{arXiv preprint arXiv:2309.01431}, 2023.

\bibitem{cui2023chatlaw}
J. Cui, Z. Li, Y. Yan, B. Chen, and L. Yuan, ‘‘Chatlaw: Open-source legal large language model with integrated external knowledge bases,’’ \textit{arXiv preprint arXiv:2306.16092}, 2023.



\bibitem{li2023muser}
Q. Li, Y. Hu, F. Yao, C. Xiao, Z. Liu, M. Sun, and W. Shen, ‘‘MUSER: A Multi-View Similar Case Retrieval Dataset,’’ in \textit{Proceedings of the 32nd ACM International Conference on Information and Knowledge Management}, 2023, pp. 5336--5340.



\bibitem{liu2023recall}
Y. Liu, L. Huang, S. Li, S. Chen, H. Zhou, F. Meng, J. Zhou, and X. Sun, ‘‘Recall: A benchmark for LLMs robustness against external counterfactual knowledge,’’ \textit{arXiv preprint arXiv:2311.08147}, 2023.


\bibitem{zhang2023retrieve}
P. Zhang, S. Xiao, Z. Liu, Z. Dou, and J.-Y. Nie, ‘‘Retrieve anything to augment large language models,’’ \textit{arXiv preprint arXiv:2310.07554}, 2023.



\bibitem{gao2022precise}
L. Gao, X. Ma, J. Lin, and J. Callan, ‘‘Precise zero-shot dense retrieval without relevance labels,’’ \textit{arXiv preprint arXiv:2212.10496}, 2022.

\bibitem{li2023lecardv2}
H. Li, Y. Shao, Y. Wu, Q. Ai, Y. Ma, and Y. Liu, ‘‘LeCaRDv2: A Large-Scale Chinese Legal Case Retrieval Dataset,’’ \textit{arXiv preprint arXiv:2310.17609}, 2023.

\bibitem{zhao2022pmc}
Z. Zhao, Q. Jin, F. Chen, T. Peng, and S. Yu, ‘‘Pmc-patients: A large-scale dataset of patient summaries and relations for benchmarking retrieval-based clinical decision support systems,’’ \textit{arXiv preprint arXiv:2202.13876}, 2022.

\bibitem{zhong2020jec}
H. Zhong, C. Xiao, C. Tu, T. Zhang, Z. Liu, and M. Sun, ‘‘JEC-QA: A legal-domain question answering dataset,’’ in \textit{Proceedings of the AAAI conference on artificial intelligence}, vol. 34, no. 05, pp. 9701--9708, 2020.

\bibitem{long2022multi}
D. Long, Q. Gao, K. Zou, G. Xu, P. Xie, R. Guo, J. Xu, G. Jiang, L. Xing, and P. Yang, ‘‘Multi-CPR: A Multi Domain Chinese Dataset for Passage Retrieval,’’ in \textit{SIGIR}, 2022, pp. 3046--3056.

\bibitem{scikit-learn}
F. Pedregosa, G. Varoquaux, A. Gramfort, V. Michel, B. Thirion, O. Grisel, M. Blondel, P. Prettenhofer, R. Weiss, V. Dubourg, J. Vanderplas, A. Passos, D. Cournapeau, M. Brucher, M. Perrot, and E. Duchesnay, ‘‘Scikit-learn: Machine learning in Python,’’ \textit{Journal of Machine Learning Research}, vol. 12, pp. 2825--2830, 2011.


\bibitem{li2025mfar}
X. Li, et al.,
“mFAR: Multi-Field Adaptive Retrieval for Semi-structured Documents,” 
in \textit{International Conference on Learning Representations (ICLR)}, 2025.


\bibitem{long2025diver}
D. Long, et al.,
“DIVER: A Multi-Stage Approach for Reasoning-intensive Information Retrieval,” 
arXiv preprint arXiv:2508.07995, 2025.
\end{thebibliography}
\end{document}